\documentclass{elsarticle}
\usepackage[utf8]{inputenc}
\usepackage{color}
\journal{Journal of \LaTeX\ Templates}
\usepackage{todonotes}
\usepackage{amsmath}
\usepackage{hyperref}
\usepackage{algorithmicx}
\usepackage{algorithm}
\usepackage{algpseudocode}

\usepackage{rotating}
\usepackage{multirow}
\usepackage{graphicx}
\usepackage{amsmath}
\usepackage[caption=false]{subfig}
\usepackage{lipsum}

\usepackage[nottoc]{tocbibind}

\begin{document}

\begin{frontmatter}
\title{AutoEn: An AutoML method based on ensembles of predefined Machine Learning pipelines for supervised Traffic Forecasting}


\author[label1,label3,mycorrespondingauthor]{Juan S. Angarita-Zapata}
\cortext[mycorrespondingauthor]{Corresponding author}
\address[label1]{Aimsun SLU, Ronda Universitat 22 B, 08007, Barcelona, Spain}
\address[label3]{DeustoTech, Faculty of Engineering, University of Deusto, Bilbao, Spain}
\ead{juan.angarita@aimsun.com}

\author[label2,label3]{Antonio D. Masegosa}
\address[label2]{IKERBASQUE, Basque Foundation for Science, Bilbao, Spain}
\ead{ad.masegosa@deusto.es}

\author[label4]{Isaac Triguero}
\address[label4]{Computer Science and Artificial Intelligence, University of Granada, Granada, Spain}
\ead{triguero@decsai.ugr.es}

\begin{abstract}
Intelligent Transportation Systems are producing tons of hardly manageable traffic data, which motivates the use of Machine Learning (ML) for data-driven applications, such as Traffic Forecasting (TF). TF is gaining relevance due to its ability to mitigate traffic congestion by forecasting future traffic states. However, TF poses one big challenge to the ML paradigm, known as the Model Selection Problem (MSP): deciding the most suitable combination of data preprocessing techniques and ML method for traffic data collected under different transportation circumstances. In this context, Automated Machine Learning (AutoML), the automation of the ML workflow from data preprocessing to model validation, arises as a promising strategy to deal with the MSP in problem domains wherein expert ML knowledge is not always an available or affordable asset, such as TF. Various AutoML frameworks have been used to approach the MSP in TF. Most are based on online optimisation processes to search for the best-performing pipeline on a given dataset. This online optimisation could be complemented with meta-learning to warm-start the search phase and/or the construction of ensembles using pipelines derived from the optimisation process. However, given the complexity of the search space and the high computational cost of tuning-evaluating pipelines generated, online optimisation is only beneficial when there is a long time to obtain the final model. Thus, we introduce AutoEn, which is a simple and efficient method for automatically generating multi-classifier ensembles from a predefined set of ML pipelines. We compare AutoEn against Auto-WEKA and Auto-sklearn, two AutoML methods commonly used in TF. Experimental results demonstrate that AutoEn can lead to better or more competitive results in the general-purpose domain and in TF. 
\end{abstract}


\end{frontmatter}



\section{Introduction}
\label{sec:introduction}

%
%
%
%

Sensing and telecommunications technologies generate vast volumes of data that motivate the use of data-driven approaches, particularly in Machine Learning (ML), to analyse this data. Like other research areas, Intelligent Transportation Systems (ITSs) announce the production of tons of hardly manageable traffic data that can be used by different applications such as traveller information systems or Traffic Forecasting (TF) schemes. Recently, TF has been gaining relevance due to its ability to deal with traffic congestion by forecasting future states of traffic measures (e.g., travel time) \cite{Angarita_etal_2019}. From a ML perspective, TF is approached through learning and approximating a mapping function from historical data to make traffic predictions when facing unseen data. 

TF poses two main challenges to the ML paradigm. First, traffic data can be collected in multiple formats (e.g., traffic-counting measures, GPS tracks) and under different transportation circumstances (e.g., urban, freeway). These characteristics influence the performance of ML methods, and choosing the most competitive method from a set of candidates brings human effort and time costs. Second, raw traffic data usually needs to be preprocessed before being analysed. Therefore, deciding the most suitable combination of data preprocessing techniques and ML method (also known as the Model Selection Problem, MSP) is a time-consuming task that demands specialised ML knowledge, which is an asset not always available. In this context, Automated Machine Learning (AutoML) is a promising path to address the MSP in TF. AutoML is an emerging area in ML that seeks to automate the ML workflow from data preprocessing to model validation \cite{automl_book2018}. Such automation provides robust AutoML methods that enable people, with either little or no specialised ML knowledge, to integrate ML solutions into data-driven processes. The latter is known as the democratisation of ML \cite{automl_book2018}, and it is aligned with the actual purpose of Artificial Intelligence: to learn and act automatically without human intervention \cite{Song2019}.

Recent literature reports a wide variety of AutoML methods, where only a few automatise the construction of complete ML pipelines \cite{Luo2016,Yao2018,Zoller2019}. In the transportation area, to the best authors’ knowledge, only a few papers have used general-purpose AutoML methods (agnostic methods w.r.t. to the problem domain where the input data comes from) for TF \cite{VLAHOGIANNI201514,Angarita2018,Angarita2020,Angarita-Zapata2020}. Those studies have used AutoML techniques that usually generate ML pipelines using an online search strategy, that is, a search strategy that takes place after the input dataset has been provided. In some cases \cite{VLAHOGIANNI201514,Angarita2018,Angarita-Zapata2020}, this online search can be purely based on optimisation approaches that test different promising combinations of pipelines structures from a predefined base of preprocessing and ML algorithms. Here the aim is to minimise or maximise a predefined performance measure (e.g., Auto-WEKA \cite{Thornton2013}). Alternatively, the study of \cite{Angarita2020} used Auto-sklearn, whose online search is complemented with meta-learning and ensemble learning \cite{Feurer2015}. This AutoML method first extracts meta-features of the input dataset at hand, providing information such as the number of instances, classes or entropy, among others. From these meta-features, meta-learning identifies suitable candidates for pipeline structures from a predefined knowledge base that stores meta-features for different datasets and pipelines that are likely to perform well on them. Then, the candidate pipelines are used for a warm-start of the optimisation approach. Lastly, the best-performing pipelines found during the optimisation search are used to build an ensemble of models.

As it can be observed from those studies in TF, the online search of AutoML generally uses optimisation as the core engine for generating and tuning pipelines. However, optimising ML pipelines is a difficult task for two reasons: (1) the complexity of the search space and (2) the evaluation cost. Regarding the search space, its complexity is given by the heterogeneity of the domains of decision variables (e.g. categorical, binary, integer). As for the evaluation cost, it is due to the need to train and validate the model on the dataset to obtain its performance. For these reasons, optimising ML pipelines usually requires a long time to provide good results, especially for big datasets. However, a lengthy optimisation process may be prone to overfitting \cite{Angarita2018,Gijsbers2019,Angarita-Zapata2020}. Although meta-learning has arisen as a promising strategy to solve some previous problems, especially the computational cost of pipeline optimisation, this approach also presents a drawback. Finding a representative set of meta-features to characterise very diverse datasets is challenging. The latter can lead to the risk that in very different ML tasks to those stored in the knowledge base, such as TF, the meta-learning component and its meta-features may suggest pipelines that do not work correctly on the dataset at hand \cite{Angarita2020}.

Considering the limitations mentioned above, the automated process of generating and testing ML pipelines in TF shows a wide margin of improvement. The latter offers the opportunity to develop new and novel AutoML approaches toward more robust and efficient strategies for the automated construction of ML pipelines, which can be better adapted to specific problem domains, such as TF. Therefore, this paper introduces a simple strategy for AutoML that does not require a pipeline optimisation process (making it more scalable and less prone to overfitting) and does not rely on extracted meta-features. The proposed approach, named AutoEn, is a simple yet effective ensemble-based AutoML method for the automatic generation of ensembles \cite{GALAR2011}, from a predefined set of pipelines composed of a sequence of preprocessing techniques plus one ML classifier. The main contributions of this paper are listed below.

\begin{itemize}
    \item To demonstrate the suitability and competitiveness of AutoEn and its simple online search strategy based on the construction of ensembles of multiple classifiers rather than in the search, construction and tuning of individual pipelines usually done by current AutoML methods in TF.
    \item To demonstrate the contributions that AutoEn brings to TF. To this end, we test AutoEn against Auto-sklearn in various multi-class TF problems previously used in the transportation literature.
    \item To test the performance and determine the competitiveness of AutoEn in the general-purpose domain (using data not related to the TF domain). This comparison is made using the first version of the so-called AutoML benchmark \cite{Gijsbers2019}, which is composed of diverse binary and multi-class supervised classification problems.
\end{itemize}

The rest of this paper is structured as follows. Section \ref{backg} presents background and related work about AutoML, emphasising methods that automatise complete ML pipelines for TF. Section \ref{automl_framework} introduces the AutoML method proposed in this paper. Later on, Section \ref{Methodology} exposes the experimental framework and Section \ref{Results} analyses the results obtained. Finally, conclusions are discussed in Section \ref{Conclusions}.

%
%
\section{Background and Related Work}
\label{backg}

\subsection{Approaches for the automatic construction of complete ML workflows}
\label{AutoML_background}

ML is the field focused on algorithms able to carry out prediction/classification tasks employing an automatic learning process without the need of being explicitly programmed by human intervention \cite{Bishop2006,Song2019}. Applications like e-mail spam and malware filtering, automatic speech recognition, and predictive maintenance in industry, among others, are built upon ML. As a common denominator, all these applications required well-designed and efficient ML pipelines that include data preprocessing as a critical factor \cite{Garcia2014,triguero19}.

Following the definition of \cite{Zoller2019}, a ML pipeline can be defined as a combination of data preprocessing methods and classifiers, with configuration of hyperparameters, that maps input data $X$ into target values $Y$ (this is known as the MSP). In this context, time, human effort, and computational capabilities are required to generate, train, and test a ML workflow because no ML pipeline can be competitive on every supervised learning task it faces. The MSP is usually approached by ML experts or by practitioners who follow a trial and error strategy, causing the success of ML to happen at a elevated costs due to the resources consumed in this task  \cite{Yao2018}


Thus, AutoML is by far a suitable strategy to deal with the MSP, while allowing to decrease human effort and bias, and computational costs by bulding ML workflows more efficiently. AutoML is an area of ML that aims at automatically finding the best combination of preprocessing techniques, ML algorithm and its hyperparameters, without being specialised in the problem domain wherein this data comes from (this is known as general-purpose AutoML) \cite{automl_book2018}. 

AutoML literature \cite{Yao2018,automl_book2018,Zoller2019} presents a variety of AutoML methods. They differ depending on which stages of a ML pipeline is automated, for example, data preprocessing \cite{Kanter2015,Katz2016,Nargesian2017}, algorithm selection and hyper-parameters \cite{Bergstra2011,Hutter2011,Thornton2013,Claesen2014}, or the entire pipeline. This paper focuses on automating ML pipelines composed of data preprocessing techniques and a classifier algorithm with their respective hyperparameter configuration. 

In the literature of the automatic construction of ML pipelines, we can usually find three main categories of AutoML methods to automatise the construction of ML pipelines. The first is purely based on optimisation approaches that test different promising combinations of algorithms from a predefined base of ML classifiers to minimise or maximise a performance measure \cite{Olson2016,Thornton2013}. Alternatively, there are AutoML methods whose online search is complemented with learning strategies like meta-learning \cite{vanschoren2019}. These techniques first extract meta-features of the input dataset (e.g., number of instances, features, classes). Then, from these meta-features, meta-learning identifies suitable candidates for pipeline structures from a predefined knowledge base that stores meta-features for different datasets and ML models that are likely to perform well on them. Then, the candidate models are typically used to warm-start an optimisation process. In addition, other AutoML methods use ensemble learning to build diverse sets of classifiers from predefined portfolios of ML algorithms \cite{Feurer2015}. These ensemble approaches have proven to be more robust than other AutoML methods, such as the case of  Auto-Gluon, which incorporates an ensemble learning strategy based on multi-layer stacking \cite{Erickson2020}.

\subsection{AutoML in Traffic Forecasting}
\label{AutoML_in_TF_background}

In the transportation area, to the best authors’ knowledge, only a few studies have used AutoML concepts in TF \cite{VLAHOGIANNI201514,Angarita2018,Angarita-Zapata2020,Angarita2020}. The first attempt was introduced by Vlahogianni et al.\cite{VLAHOGIANNI201514}. They proposed a meta-modelling technique that automatically recommends a classifier for traffic prediction. This method was based on surrogate modelling and a genetic algorithm with an island model. The authors automatised the algorithm selection and the hyper-parameter setting, without including the preprocessing stage, as this paper does. The study of Vlahogianni et al.\cite{VLAHOGIANNI201514} can be considered the first incursion of AutoML principles in the TF domain.

After that, Angarita et al. in \cite{Angarita2018,Angarita-Zapata2020,Angarita2020} brought AutoML state-fo-the-art concepts to the TF area. In these three studies, the authors determined the extent to which general-purpose AutoML can be competitive against ad hoc methods in domain-related problems. Specifically, they researched whether it is valuable to use general-purpose AutoML already available in the literature (they used Auto-WEKA and Auto-sklearn) or develop domain-related AutoML for ITSs, using TF as a case study.

In the case of \cite{Angarita-Zapata2020} and \cite{Angarita2018}, the authors used Auto-WEKA, an AutoML method that applies sequential model-based Bayesian optimisation to find optimal ML pipelines for TF. Both papers tested the performance of Auto-WEKA against the classic model selection approach, which consists of selecting the best of a set of algorithms to predict traffic by trial and error. Lastly, Angarita et al. \cite{Angarita2020} used Auto-sklearn version one, a state-of-the-art AutoML method whose search strategy of pipelines uses Bayesian optimisation, meta-learning and ensemble learning. The authors tested this method in multi-class imbalanced classification problems for different time horizons and for freeway and urban environments.

As a common factor, the core of pipeline search strategies used by the AutoML methods tested in \cite{VLAHOGIANNI201514,Angarita2018,Angarita-Zapata2020,Angarita2020} is focused on generating and fine-tuning individual pipelines. Nonetheless, this online optimisation of pipelines is a demanding process because 1) it involves complex search spaces; 2) the evaluation of the objective function usually is computationally expensive; and 3) to establish a priori the best time budget for the optimisation process is a difficult task since in complex or big datasets collecting good pipelines can require a long time, while in small or simpler datasets an excessive time budget is prone to overfitting \cite{Angarita2018,Angarita-Zapata2020}. Although Auto-sklearn combines meta-learning with optimisation to reduce the impact of some of these issues, it has a drawback. Defining a set of meta-features that characterises very diverse datasets is difficult without solid evidence to guide this design process. In this way, there is a risk that in very different supervised learning tasks to those included in the meta-knowledge base, such as the case of TF, the meta-learning component may suggest pipelines that are not competitive to warm-start the optimisation process \cite{Angarita2020}.

Those in-depth analyses of general-purpose AutoML in TF allowed identifying a set of strengths and weaknesses of AutoML when dealing with supervised traffic prediction. Therefore, in this paper, we introduce AutoEn, a new and novel AutoML method that overcomes the drawbacks (related to optimisation and meta-learning) identified by\cite{Angarita2018,Angarita-Zapata2020,Angarita2020}. AutoEn enhances the AutoML state-of-the-art in TF by automatically generating ensembles during the online search phase of AutoML. Thus, rather than focusing on single pipelines, we approach the finding of solid ensembles of multiple pipelines, which is a more straightforward and efficient strategy to deal with the issues mentioned above. The latter enables us to improve the performance of AutoML in TF, which can also benefit the performance of AutoML in the general-purpose domain.  

%
%
\section{AutoEn: An AutoML method based on Ensemble learning}
\label{proposed_automl_framework}

\subsection{Motivation}
\label{automl_motivation}

As we stated before, the core of the existing AutoML methods used in TF is the optimisation of individual ML pipelines, which could be enhanced with a preliminary stage based on meta-learning. However, the current solutions suffer from several issues that motivate the design of AutoEn. 

\begin{itemize}
    
    \item \textbf{Optimisation - Overfitting issues}: Optimising pipelines is supposed to require long time budgets to find competitive solutions. However, previous research has corroborated that longer execution times may cause overfitting issues due to hyperparameters tuning \cite{Angarita2018,Gijsbers2019,Angarita-Zapata2020}. Therefore, higher execution times do not always lead to better results as we could expect. In addition, for medium- and small-size datasets, competitive pipelines can be quickly found without the need to allocate long time budgets for the optimisation search. Thus, optimising pipelines can be expensive when the goal is to make good enough predictions.
    
    \item \textbf{Optimisation - Reduced scalability}: Although long time budgets of optimisation may be prone to return pipelines with overfitting issues, other times the results can be promising. Particularly, when optimisation deals with small- or medium-size datasets, many diverse pipeline structures can be generated, tuned and tested because the complexity of the learning task at hand is influenced by its data size. On the other hand, the latter may stop happening as the data size grows. In this scenario, it is harder to tune and test multiple pipelines, as the optimisation becomes expensive, and the set of candidate pipelines could decrease, which ends up affecting the performance of the final solutions.

    \item \textbf{Meta-learning - Representativeness of Meta-features}: Meta-learning can potentially reduce the cost of the optimisation process. Nevertheless, there is a risk that the meta-features can only describe the learning tasks in the meta-knowledge base. The latter is understandable because it is difficult to find a set of meta-features good enough to characterise very diverse datasets. Therefore, there is a risk that for supervised learning tasks not similar to the ones stored in the meta-knowledge base, the meta-learning component recommends pipelines that do not perform as well as expected in specific supervised learning tasks \cite{Angarita2020}. Thus, the optimisation process could be warm-started with pipelines that are not competitive on the input data.
    
\end{itemize}

Keeping in mind the motivations presented above, we want to conceive a competitive and straightforward AutoML method that adjusts to the complexity of the data without the need to define a time budget for its execution. This means that for small- and medium-size datasets, the method can find quick solutions; conversely, for large datasets, the method may take longer time to find competitive pipelines. The aforementioned condition could be satisfied by not only suggesting single pipelines whose performance could vary drastically from one learning task to another. In this sense, we propose an AutoML method based on the search and construction of ensembles of multi-classifiers that do not commit to a single ML workflow. Accomplishing such a goal requires the ensemble contains models that individually are different in nature and strong in making individual predictions. Therefore, we introduce AutoEn, an AutoML method based on ensemble learning for combining high-performance instantiations of different pipelines.

\subsection{AutoEn design and workflow}
\label{automl_design_workflow}

\begin{figure}[h]
\caption{General workflow of AutoEn that is composed of two main blocks: 1) a set of predefined pipelines trained on different learning tasks and 2) an ensemble component that generates a multi-classifier system, from the set of pipelines, when a new dataset comes in.}
\label{automl_workflow_overview}
\centering
\includegraphics[scale=0.25]{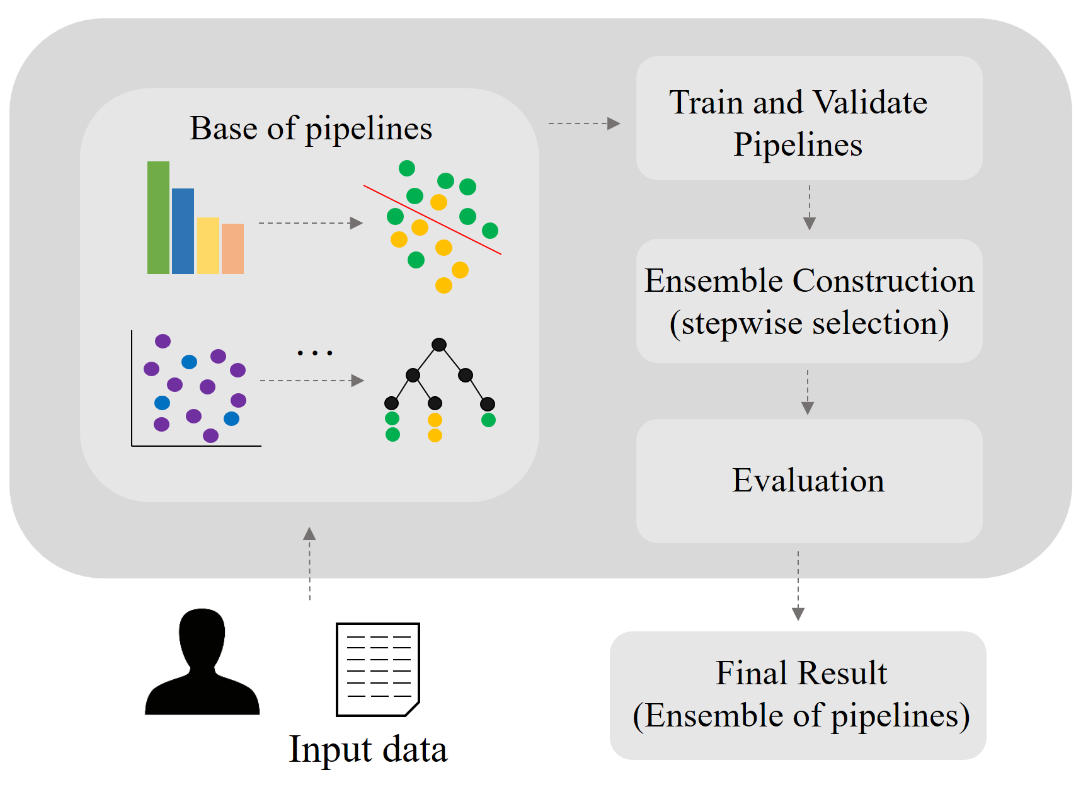}
\end{figure}

Figure \ref{automl_workflow_overview} introduces the architecture of AutoEn. This AutoML replaces the online search and optimisation of individual pipelines with the automated generation of ensembles. Similarly to Auto-sklearn, it has in its inner structure a base of diverse pipelines that were trained on different learning tasks during the construction of AutoEn (the off-line phase of AutoML). The pipelines within this base consist of a sequence of data preprocessing techniques and a classifier algorithm. The underlying idea of using a set of predefined pipelines is offering ML workflows different in nature for every learning task,  which are less prone to overfitting issues. Unlike Auto-sklearn, we will not use meta-features to decide what pipelines are part or not of the ensemble construction.

A meta-learning approach should be able to assist us in finding the most promising pipelines to be part of the ensemble. However, we observed in \cite{Angarita2020} that this idea can only correctly work under a good enough set of meta-features able to characterise very different datasets, which unfortunately is not always the case. To avoid relying on meta-features, when new input data comes in AutoEn, it assesses the performance of the base of pipelines on a validation set. The pipelines can or cannot be part of the ensemble construction depending on their validation errors. Thus, we avoid the drawbacks of meta-learning when facing very diverse learning tasks.

\begin{figure}[h]
\caption{Automated selection and construction of the ensemble based on the set of pipelines .}
\label{automl_framework}
\centering
\includegraphics[scale=0.3]{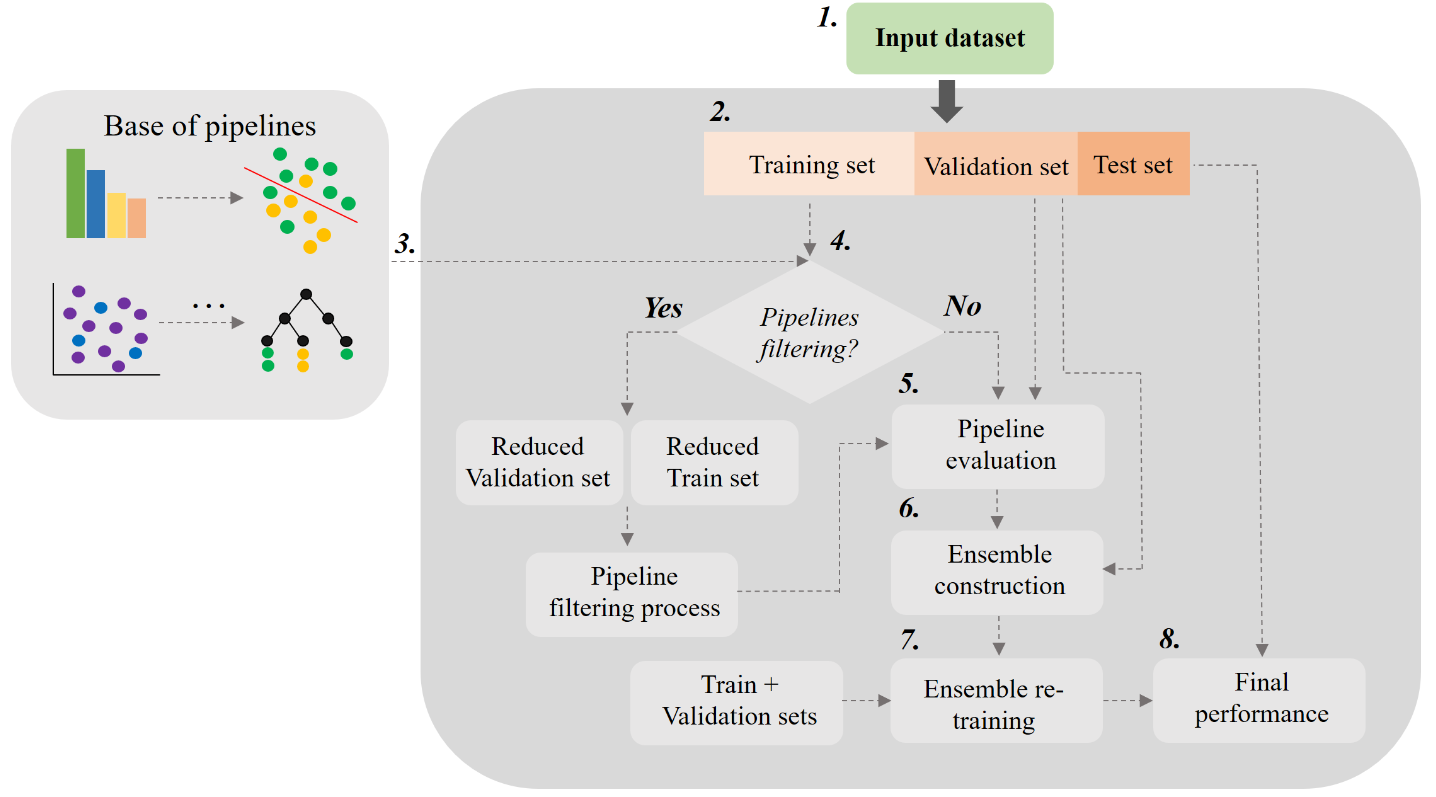}
\end{figure}

Details about the automatic selection and construction of the ensemble based on these pipelines is presented in Figure \ref{automl_framework}. In step (1), it receives a new and unseen dataset that later in step (2) is split into train, validation and test sets. Then, in step (3), we retrieve the set of available ML pipelines.

Having the list of optimised pipelines at hand, in step (4), it is possible to filter them to improve the computational efficiency or not. The latter decision implies either directly training the complete list of pipelines on the training set ("No" path) or reducing the list to have a smaller set ("Yes" path) to enhance the computational efficiency. The former option is the default mode of our AutoML method, whilst the latter implies selecting a sample of the training set (dashed line in Figure \ref{automl_framework}), which is divided into training and validation set (reduced train and validation set in Figure \ref{automl_framework}). The aim is to filter the pipelines and choose the fastest ones.

After selecting the complete list of pipelines or a reduced version, the next step is training the available pipelines in the complete train set and ranking them according to their errors in the validation set (step 5). After that, we build an ensemble from this set of trained pipelines (step 6). Finally, we implemented the ensemble selection approach introduced by \cite{Caruana2004}. The latter is a greedy procedure that starts from an empty ensemble and then iteratively adds the model that maximises ensemble performance in the same validation set in which the pipelines were previously assessed. 

For our AutoEn, we initialise the ensemble with one element, the pipeline with the highest validation set performance. Then, from the one-element ensemble, we start to append new pipelines, with uniform weights, to the one-element ensemble until reaching a predefined ensemble size. It is essential to note that the inclusion of new pipelines in the ensemble allows repetitions; this means one high-performance pipeline can appear multiple times in the final ensemble. Finally, in step (7), the pipelines that composed the ensemble are re-trained on training+validation partitions to make classification on the unseen test set finally. For this final step, the pipelines that appear multiple times in the ensemble are only re-trained once; thus, we can decrease the computational cost of this final step.

%
%
\section{Experimental Framework}
\label{Methodology}

This paper seeks to answer the question of whether it is possible to conceive an AutoML framework based on multi-classifier ensembles and able to make good enough predictions in supervised classification problems, specifically for the case of supervised TF. We test AutoEn in the general-purpose domain and the TF problems to accomplish such a purpose. 

Regarding the general-purpose area, we compare AutoEn against Auto-sklearn and Auto-WEKA, two well-established AutoML methods based on pure optimisation and meta-learning with optimisation, which have been used in TF literature previously. We evaluate the performance of AutoEn versus the results of Auto-sklearn and Auto-WEKA reported in the AutoML benchmark published by \cite{Gijsbers2019}. This benchmark is an open-source AutoML framework that uses public datasets to compare AutoML methods thoroughly. 

Regarding the TF domain, we demonstrate that AutoEn enhances the performance of AutoML in TF. To this end, we test AutoEn against Auto-sklearn in various multi-class TF problems that were previously used in \cite{Angarita2020}. The reason to only consider Auto-sklearn is that this AutoML method overcomes the limitations of AutoML techniques based only on optimisation, such as in the case of Auto-WEKA.

The rest of this section presents the elements related to the two experiments designed to test and assess the competitiveness of AutoEn. First, we provide details of the datasets chosen for the experimentation. Then, we introduce the measures employed to evaluate the performance of the methods. Finally, we present the methods used for comparison.

\subsection{Supervised learning problems}
\label{exp_AutoEn_general-purpose_domain}

\subsubsection{General-purpose datasets}

We use 16 datasets of binary classification problems and 12 multi-class problems from the AutoML benchmark developed by \cite{Gijsbers2019}. The total of 28 datasets are not related to TF and were used in previous AutoML papers \cite{Thornton2013}, AutoML competitions \cite{Isabelle2016} and ML benchmarks \cite{bischl2017}. The datasets vary in the number of samples and features by orders of magnitude and the occurrence of numeric features, categorical features and missing values. The current list of datasets is available on OpenML \cite{Vanschoren2014} 

Tables \ref{binary-datasets} and \ref{multicass-datasets} present a summary of binary and multi-class datasets used for this experimentation. They show the number of features and instances (Feat. - Instan.), the number of numeric and nominal features (Num. Feat. and Nom. Feat), and the number of missing values (Missing Val.) on each dataset. 

\begin{table}[t!]
\centering
\caption{Binary datasets}
\label{binary-datasets}
\resizebox{\columnwidth}{!}{%
\begin{tabular}{ccccc}
\hline
\textbf{dataset} & \textbf{Feat. - Instan.} & \textbf{Num. Feat.} & \textbf{Num. Feat.} & \textbf{Missing Val.} \\ \hline
adult & 15 - 48842 & 6 & 9 & 6465 \\ 
amazon\_employ & 10 - 32769 & 0 & 10 & 0 \\ 
albert & 79 - 425240 & 26 & 53 & 2734000 \\ 
apsfailure & 171 - 76000 & 170 & 1 & 1078695 \\ 
bank-mark & 17 - 45211 & 7 & 10 & 0 \\ 
blood-transf & 5 - 748 & 4 & 1 & 0 \\ 
christine & 1637 - 5418 & 1599 & 38 & 0 \\ 
credit-g & 21 - 1000 & 7 & 14 & 0 \\ 
higgs & 29 - 98050 & 28 & 1 & 9 \\ 
jasmine & 145 - 2984 & 8 & 137 & 0 \\ 
kc1 & 22 - 2109 & 21 & 1 & 0 \\ 
kr-vs-kp & 37 - 3196 & 0 & 37 & 0 \\ 
nomao & 119 - 34465 & 89 & 30 & 0 \\
numerai28.6 & 22 - 92320 & 21 & 1 & 0 \\ 
phoneme & 6 - 5404 & 5 & 1 & 0 \\ 
sylvine & 21 - 5124 & 20 & 1 & 0 \\ \hline
\end{tabular}}
\end{table}

\begin{table}[t!]
\centering
\caption{Multi-class datasets}
\label{multicass-datasets}
\resizebox{\columnwidth}{!}{%
\begin{tabular}{ccccc}
\hline
\textbf{dataset} & \textbf{Feat. - Instan} & \textbf{Num. Feat.} & \textbf{Nom. Feat.} & \textbf{Classes} \\ \hline
car & 7 - 1728 & 0 & 7 & 4 \\ 
cnae-9 & 857 - 1080 & 856 & 0 & 9 \\
connect-4 & 43 - 67557 & 0 & 43 & 3 \\ 
dilbert & 2001 - 10000 & 2000 & 1 & 5 \\ 
fashion-mnist & 785 - 70000 & 784 & 1 & 10 \\ 
jannis & 55 - 83733 & 54 & 1 & 4 \\
jungle & 7 - 44819 & 6 & 1 & 3 \\
mfeat-factors & 217 - 2000 & 216 & 1 & 10 \\ 
segment & 20 - 2310 & 19 & 1 & 7 \\
shuttle & 10 - 58000 & 9 & 1 & 7 \\ 
vehicle & 19 - 846 & 18 & 1 & 4 \\ 
volkert & 181 - 58310 & 180 & 1 & 10 \\ \hline
\end{tabular}}
\end{table}

\subsubsection{Traffic Forecasting datasets}

We considered freeway and urban scenarios for these datasets. In the freeway case, the data was taken from Caltrans Performance Measurement System \footnote{http://pems.dot.ca.gov}; meanwhile, for the urban enviroment the data was retrieved from Madrid Open Data Portal \footnote{https://datos.madrid.es/portal/site/egob/}. In freeway and urban cases, the traffic measure was three months of speed in aggregation times of 5 and 15 minutes, respectively. 

All datasets were enriched with Calendar Data (Day of the week and Minute of the day). In addition to Calendar dDta (CD), freeway and urban datasets can be composed of only temporal traffic data (Temporal traffic data, T) from the target location or traffic data from the target location and four downstream locations (Temporal and Spatial traffic data, TS). More details of these datasets can be consulted in \cite{Angarita2020, Angarita-Zapata2020}.

The target output of all datasets is modelled as multi-class classification problem where the aim is to predict the Level of Service (LoS). LoS is a traffic measure topically used to describe the quality of traffic flows. That categorization of traffic is done using categorical levels, which range from A to E in a gradual way\footnote{Category \textit{A} indicates light to moderate traffic, whereas a category \textit{E} represents extended traffic delays} \cite{Skycomp2009}. In the freeway case, the time horizons of predictions are 5, 15, 30, 45, and 60 minutes using a data granularity of 5 minutes. For the urban datasets, the forecasting time steps are 15, 30, 45, and 60 minutes with a data granularity of 15 minutes.
    
Table \ref{Table:data_sets} presents a summary of the datasets described above. The description includes the number of instances, features, and the number of classes of each dataset.

\begin{table}[tbp]
\centering
\caption{Traffic Forecasting datasets}
\label{Table:data_sets}
\resizebox{\columnwidth}{!}{%
\begin{tabular}{cccccc}
\hline
\textbf{Environment} & \textbf{datasets} & \textbf{Num. Instan.} & \textbf{Featurs} & \textbf{Classes} \\ \hline
\multirow{2}{*}{Freeway} & \begin{tabular}[c]{@{}c@{}}T+CD in time horizons of \\ 5, 15, 30, 45, 60 minutes\end{tabular} & {[}10927 - 9906{]} & 13 & \begin{tabular}[c]{@{}c@{}} 4 \end{tabular} \\\\
 & \begin{tabular}[c]{@{}c@{}}TS+CD in time horizons of \\ 5, 15, 30, 45, 60 minutes\end{tabular} & {[}10927 - 9906{]} & 28 & \begin{tabular}[c]{@{}c@{}} 3 \end{tabular} \\ \hline
\multirow{2}{*}{Urban} & \begin{tabular}[c]{@{}c@{}}T+CD in time horizons of \\ 15, 30, 45, 60 minutes\end{tabular} & {[}2684-2634{]} & 13 & \begin{tabular}[c]{@{}c@{}} 3 \end{tabular} \\\\
 & \begin{tabular}[c]{@{}c@{}}TS+CD in time horizons of \\ 15, 30, 45, 60 minutes\end{tabular} & {[}2684-2634{]} & 28 & \begin{tabular}[c]{@{}c@{}} 3 \end{tabular} \\ \hline
\end{tabular}}
\end{table}

\subsection{Performance metrics and Hardware choice}
\label{Performance measures}

\subsubsection{General-purpose domain}
\label{Performance measures_general_domain}

\begin{itemize}
    \item \textbf{Metrics:} For the results of this paper, we follow the same experimental set-up proposed in the AutoML benchmark to make fair comparisons, and therefore, we use the same performance measures. The area under the receiver operator curve (roc\_auc\_score) is used for binary classification problems and log loss\_score for multi-class problems. Besides, both measures are averages of ten-fold cross-validation.
    \item \textbf{Hardware choice and resource specifications:} We opted to use a cluster based on Centos 7.6 with kernel 3.10.0, and usig QuadCore Intel Xeon processors, with 16 cores and 16 GB of RAM each. Here, we highlight that despite the multi-core character of the CPU, the implementation of AutoEn is single-core.
    \item \textbf{Computational time:} To measure how much time AutEn takes since new input data comes in until it makes final predictions, we measure the execution time in seconds extracted from the system clock. Concretely, we start to measure this time when AutoEn receives the input data from making the data partitions until it outputs the final classification of the ensemble built during the online search of the method.
    \item \textbf{Statistical tests:} We used non-parametric statistical tests to assess the differences in performance of the methods. Two statistical tests are considred following the guidelines proposed in \cite{GARCIA2010}. First, Friedman's test for multiple comparisons is applied to check whether there are differences among the methods. Then, Holm's test is used to check whether the differences in the Friedman ranking are statistically significant or not.
\end{itemize}

\subsubsection{Traffic Forecasting}

\begin{itemize}
    \item\textbf{Metrics:} We use for this case study the same experimental framework for multi-class problems proposed in Section \ref{Performance measures_general_domain}. This means that we assess the performance of results using loss\_score averaged over ten-fold cross-validation per dataset.
    \item \textbf{Hardware choice and resource specifications:} We used the same cluster considered for the experimentation carried out in the general-purpose domain.
    \item \textbf{Computational time:} For the case of TF, we did not measure the computational time because of the small size of the datasets used, so all the methods will have very similar computational times it was measured.
    \item \textbf{Statistical tests:} We also made use of the same two non-parametric statistical tests to assess the differences in the performance of the methods: Friedman's and  Holm's tests.
\end{itemize}

\subsection{Competitors and baseline}
\label{Competitors and Baseline}

For the experimentation carried out in this paper, AutoEn has used the pipelines list stored in the first version of Auto-sklearn \cite{Feurer2015}. These pipelines were generated and tuned using sequential mode-based optimisation \cite{Hutter2011} using a search space of 15 classifiers and 18 preprocessing techniques, all implemented in scikit-learn ML library. The classifier can be categorised in linear models, support vector machines, discriminant analysis, nearest neighbors, naive Bayes, decision trees, and ensembles. Data preprocessing techniques include rescaling, imputation of missing values, one-hot encoding, feature selection, kernel approximation, feature clustering, and polynomial feature expansion. The interested reader can consult \cite{Falkne2018} in order to know more details about these classifiers and preprocessing techniques.

\subsubsection{General-purpose domain}

As mentioned above, the referenced AutoML benchmark makes comparisons of four AutoML state-of-the-art methods: H2O, Auto-WEKA, Auto-sklearn, and TPOT. In this paper, we make comparisons against the same AutoML methods (Auto-WEKA and Auto-sklearn), which have been used previously to approach supervised TF problems \cite{Angarita2018,Angarita2020,Angarita-Zapata2020}. Additionally, we include the same baselines methods of the referenced AutoML benchmark. They are a constant predictor, which always predicts the class prior (CnstPrd), an untuned Random Forest (RF), and a tuned Random Forest (tuned\_RF). We also include the winner approach (BestV\_ML) of the experimentation carried out with Auto-sklearn version one by \cite{Angarita2020}.

Auto-sklearn \cite{Feurer2015} and Auto-WEKA \cite{Thornton2013} were deployed with their default hyperparameter values and search spaces, since most users will use them in this way, and their time budget per fold was 1 and 4 hours. RF uses scikit-learn 0.20 default hyperparameters, and tuned\_RF is built with 2000 estimators. For AutoEn and AutoEn\_economy (AutoEn\_ec), their main parameters are shown in Table \ref{autoen_hyperparameters}.

\begin{table}[h]
\centering
\caption{Initial hyperparameters of AutoEn for its two operative modes}
\label{autoen_hyperparameters}
\resizebox{0.65\columnwidth}{!}{%
\begin{tabular}{ccc}
\hline
\textbf{Hyperparameters} & \textbf{AutoEn} & \textbf{AutoEn\_ec} \\ \hline
\\Ensemble size & 50 & 50 \\\hline
Data partition per fold & \begin{tabular}[c]{@{}c@{}}train: 60\%,\\ validation: 20\%,\\  test: 20\%\end{tabular} & \begin{tabular}[c]{@{}c@{}}train: 60\%,\\ validation: 20\%,\\  test: 20\%\end{tabular} \\\hline
\begin{tabular}[c]{@{}c@{}}Data partition to \\ filter pipelines\end{tabular} & - & \begin{tabular}[c]{@{}c@{}}10\% of the \\ original train set\end{tabular} \\\hline 
\begin{tabular}[c]{@{}c@{}}Time left for a pipeline \\ to be trained in the pipelines\\  filtering phase\end{tabular} & - & 36 seconds \\\hline
\end{tabular}}
\end{table}

\subsubsection{Traffic Forecasting}

AutoEn used again the pipelines list stored in Auto-sklearn's knowledge base (version one) for the experimentation carried out in the case study. Besides, we did not consider AutoEn\_ec due to the size of the datasets considered. Freeway datasets are around 10.000 instances, and urban datasets have approximately 2.500 instances. 

In the case of AutoML competitors, we have made comparisons against Auto-sklearn with its default hyperparameter values using three execution times (ET): 15, 60, and 150 minutes. Each ET is considered an individual AutoML competitor with an allocated time to find competitive ML pipelines. Additionally, as baseline methods, we use a tuned RF built with 2000 estimators and BestV\_ML, which was the winner technique in the experimentation carried out with Auto-sklearn version one in \cite{Angarita2020}.

%
%

\section{Analysis of Results}
\label{Results}

\subsection{Results in the general-purpose domain}

This section analyses the results obtained from different angles without being focused on TF problems. The latter will allows us to show how our AutoML proposal can contribute not only to the TF, but also to the general-purpose area where various learning problems can be found. Specifically, our aims are:

\begin{itemize}

    \item To compare the competitiveness and the significance of AutoEn performance with respect to AutoML competitors and baselines in binary and multi-class learning tasks.
    \item To investigate the main benefits of an AutoML method without allocating any time budget when dealing with classification datasets of different sizes.
    \item To contrast the differences in performance and runtime between AutoEn and its economy mode.

\end{itemize}

Table \ref{Table:1-4_hours_results} shows the mean $roc\_auc\_score$ and $log\_loss\_score$ values of our AutoML method in its default (AutoEn) and economy modes (AutoEn\_ec), the AutoML competitors (AutoSkl\_1h, AutoSkl\_4h, AutoW\_1h, AutoW\_4h), and the baseline methods on the test phase. The best result for each dataset is highlighted in bold-face. Note that for binary datasets, the higher the performance value (roc\_auc), the better; for multi-class datasets, the lower the loss, the better.

\begin{table}[h]
\centering
\caption{Mean $roc\_auc$ (binary problems) and $log\_loss$ (multi-class problems) values obtained by AutoEn, AutoEn\_ec, the AutoML competitors and the baseline methods. Values highlighted in bold are the highest performance obtained by any of the methods on every dataset.}
\label{Table:1-4_hours_results}
\resizebox{\columnwidth}{!}{%
\begin{tabular}{clcccccccccc}
\hline
Type & \multicolumn{1}{c}{Dataset} & AutoSkl\_1h & AutoSkl\_4h & AutoW\_1h & AutoW\_4h & ConstPrd & RF & tuned\_RF & BestV\_ML & \textbf{AutoEn} & \textbf{AutoEn\_ec} \\ \hline
\multirow{16}{*}{Binary} & adult & \textbf{0.930} & \textbf{0.930} & 0.908 & 0.909 & 0.500 & 0.909 & 0.909 & 0.917 & 0.920 & 0.920 \\
 & amazon\_employee & 0.856 & 0.849 & 0.809 & 0.820 & 0.500 & \textbf{0.864} & 0.863 & 0.859 & \textbf{0.864} & 0.863 \\
 & albert & \textbf{0.748} & \textbf{0.748} & 0.724 & 0.724 & 0.500 & 0.738 & 0.738 & 0.739 & 0.702 & 0.704 \\
 & apsfailure & 0.991 & \textbf{0.992} & 0.965 & 0.984 & 0.500 & 0.991 & 0.991 & 0.990 & 0.991 & 0.989 \\
 & bank-marketing & \textbf{0.937} & \textbf{0.937} & 0.827 & 0.909 & 0.500 & 0.931 & 0.931 & 0.931 & 0.934 & 0.935 \\
 & blood-transfusion & 0.757 & \textbf{0.763} & 0.741 & 0.742 & 0.500 & 0.686 & 0.689 & 0.730 & 0.741 & 0.735 \\
 & christine & 0.830 & \textbf{0.831} & 0.802 & 0.809 & 0.500 & 0.806 & 0.810 & 0.816 & 0.825 & 0.811 \\
 & credit-g & 0.783 & 0.782 & 0.753 & 0.744 & 0.500 & 0.795 & \textbf{0.796} & 0.781 & 0.786 & 0.795 \\
 & higgs & 0.793 & \textbf{0.809} & 0.677 & 0.757 & 0.500 & 0.803 & 0.803 & 0.798 & 0.807 & 0.807 \\
 & jasmine & 0.884 & 0.883 & 0.861 & 0.865 & 0.500 & 0.888 & \textbf{0.889} & 0.878 & 0.881 & 0.883 \\
 & kc1 & 0.843 & 0.839 & 0.814 & 0.818 & 0.500 & 0.836 & 0.842 & 0.840 & \textbf{0.852} & 0.844 \\
 & kr-vs-kp & \textbf{1.000} & \textbf{1.000} & 0.976 & 0.979 & 0.500 & 0.999 & \textbf{1.000} & \textbf{1.000} & 0.998 & 0.998 \\
 & nomao & \textbf{0.996} & \textbf{0.996} & 0.984 & 0.982 & 0.500 & 0.995 & 0.995 & 0.995 & \textbf{0.996} & 0.969 \\
 & numerai28.6 & 0.529 & 0.530 & 0.520 & 0.528 & 0.500 & 0.520 & 0.521 & 0.529 & \textbf{0.532} & 0.530 \\
 & phoneme & 0.963 & 0.962 & 0.957 & 0.965 & 0.500 & 0.965 & 0.966 & 0.970 & \textbf{0.972} & 0.970 \\
 & sylvine & 0.990 & \textbf{0.991} & 0.975 & 0.977 & 0.500 & 0.983 & 0.984 & 0.989 & 0.989 & 0.990 \\ \hline
\multirow{12}{*}{Multi-class} & car & \textbf{0.010} & \textbf{0.010} & 0.243 & 0.122 & 0.836 & 0.144 & 0.047 & 0.105 & 0.065 & 0.070 \\
 & cnae-9 & 0.171 & \textbf{0.168} & 0.873 & 1.173 & 2.197 & 0.301 & 0.297 & 0.205 & 0.178 & 0.182 \\
 & connect-4 & 0.426 & \textbf{0.387} & 0.741 & 1.427 & 0.845 & 0.495 & 0.478 & 0.483 & 0.460 & 0.461 \\
 & dilbert & 0.097 & 0.063 & 1.787 & 1.009 & 1.609 & 0.328 & 0.329 & \textbf{0.033} & \textbf{0.033} & 0.034 \\
 & fashion-mnist & 0.354 & 0.358 & 0.581 & 0.902 & 2.303 & 0.361 & 0.362 & 0.345 & \textbf{0.314} & 0.315 \\
 & jannis & 0.705 & \textbf{0.685} & 6.271 & 1.885 & 1.109 & 0.728 & 0.729 & 0.710 & 0.702 & 0.701 \\
 & jungle & 0.234 & 0.223 & 1.559 & 2.695 & 0.935 & 0.438 & 0.402 & 0.216 & \textbf{0.215} & \textbf{0.215} \\
 & mfeat-factors & 0.099 & 0.093 & 0.627 & 0.656 & 2.303 & 0.234 & 0.201 & 0.160 & 0.093 & \textbf{0.088} \\
 & segment & \textbf{0.060} & 0.063 & 0.501 & 0.427 & 1.946 & 0.084 & 0.069 & 0.074 & 0.061 & 0.069 \\
 & shuttle & 0.001 & \textbf{0.000} & 0.015 & 0.015 & 0.666 & 0.001 & 0.001 & \textbf{0.000} & \textbf{0.000} & 0.001 \\
 & vehicle & 0.395 & 0.379 & 2.105 & 5.560 & 1.386 & 0.497 & 0.486 & 0.352 & 0.341 & \textbf{0.332} \\
 & volkert & 0.945 & 0.925 & 1.110 & 8.329 & 2.053 & 0.980 & 0.979 & 0.925 & \textbf{0.858} & \textbf{0.858} \\ \cline{1-12} 
\end{tabular}}
\end{table}

Observing Table \ref{Table:1-4_hours_results}, we can point out the following:

\begin{itemize}
    \item As general overview, in binary datasets there are ties of at least 2 methods in 5 datasets: \textit{adult}, \textit{albert}, \textit{bank\_marketing}, \textit{kr-vs-kp}, and \textit{nomao}. With respect to multi-class datasets, there are 5 ties of at least 2 methods in \textit{car}, \textit{dilbert}, \textit{dilbert}, \textit{jungle} and \textit{volkert} datasets. Summarising , there is no AutoML method that consistently outperforms all AutoML competitors on binary and multi-class; and AutoML scores are relatively close to the performance of tuned\_RF. Additionally, any of the AutoML methods outperforms the tuned\_RF on the complete list of datasets.
    
    \item Another interesting aspect of results in Table \ref{Table:1-4_hours_results} is Auto-sklearn and Auto-WEKA results are quite similar under the time budgets of 1 and 4 hours per fold. The latter means that the optimisation process carries out by those 2 methods only brings slight score improvements. Particularly for binary-datasets, Auto-sklearn only achieves improvements with a longer execution time in \textit{apsfailure}, \textit{blood-transfusion}, \textit{christine}, \textit{numerai28.6} and \textit{sylvine}. Those improvements range from 0.001 to 0.007. In \textit{amazon\_employee} , \textit{credit-g}, \textit{jasmine}, \textit{kc-1} and \textit{phoneme}, Auto-sklearn obtains worse performance with 4 hours execution time than the performance it obtains with only 1 hour. For the case of multi-class datasets, although the improvements are greater, they do not reach at least 10\% of enhancement; datasets \textit{cnae-9}, \textit{connect-4}, \textit{dilbert}, \textit{jannis}, \textit{vehicle} and \textit{volkert} exemplified the aforementioned situation. As well as binary datasets, sometimes Auto-sklearn performance decreases with longer execution times in \textit{fashion-mnist} and \textit{segment} datasets. Such worsening under longer time budgets allocated for the its optimisation could be due to overfitting issues such as it has been corroborated by previous research \cite{Angarita2020,Gijsbers2019}. 
    
    \item For the case of Auto-WEKA, the behaviour of improvements from 1 to 4 hours are quite similar with respect to Auto-sklearn. In binary and multi-class datasets there are supervised problems wherein overfitting also seems to affect the performance of Auto-WEKA (\textit{adult}, \textit{albert}, \textit{blood-transfusion}, \textit{jasmine}, \textit{kc1}, \textit{kr-vs-kp},\textit{mefeat-factors}, \textit{suttle} datasets). In only some datasets there are significant improvements with regard to the shortest execution time (\textit{amazon-employee}, \textit{apsfailure}, \textit{bank-marketing}, and \textit{higgs} dataset). Moreover, there is a considerable worsening in performance when Auto-WEKA receives a longer execution time in some multi-class problems: \textit{cnae-9}, \textit{connect-4}, \textit{jungle}, \textit{vehicle} and \textit{volkert}.
    
\end{itemize}

To assess whether the differences in performance observed in Table \ref{Table:1-4_hours_results} are significant or not, we used non-parametric statistical tests (Friedman's test and Holm post-hoc test). Table \ref{tab:fiendman_testall_Holms_binary} exposes the test outcomes for binary datasets and again $p$-values lower than 0.05 are shown in bold. In this case, AutoSkl\_4h is the method in the first position of the ranking. However, it is interesting to note AutoSkl\_4h is only statistical better than AutoW\_1h and Auto\_4h, which is the AutoML with a search strategy only based on optimisation. The rest of the methods, including RF, tuned\_TF and BestV\_ML, have better performance than the latter AutoML approach.


\begin{table}[h]
\centering
\caption{Binary datasets: Friedman's average ranking and $p$-values obtained through Holm post-hoc test using AutoSkl\_4h as control method}
\label{tab:fiendman_testall_Holms_binary}
\resizebox{0.45\columnwidth}{!}{%
\begin{tabular}{ccc}
\hline
\textbf{Methods} & \textbf{Av. Ranking} & \textbf{$p$-values} \\ \hline
AutoSkl\_4h & 3 & - \\ 
AutoSkl\_1h & 3.4062 & 1 \\ 
\textbf{AutoEn} & 3.5938 & 1 \\ 
\textbf{AutoEn\_ec} & 4.4062 & 0.4391 \\ \
tuned\_RF & 4.5938 & 0.3990 \\ 
BestV\_ML & 5 & 0.1943 \\
RF & 5.4062 & 0.0776 \\ 
AutoW\_4h & 7.25 & \textbf{0.0008} \\ 
AutoW\_1h & 8.3438 & \textbf{0} \\ \hline
\end{tabular}}
\end{table}

Table \ref{tab:fiendman_testall_Holms_multiclass} summarises test results for multi-class datasets and significant differences with $p$-values lower than 0.05 are highlighted in bold. AutoEn is the method in the first position of Friedman's average ranking and such as binary problems, there are no significant statistical differences among AutoEn, Auto-sklearn and BestV\_ML methods. Moreover, in multi-class datasets, AutoEn is better than Auto-WEKA and the two variants of RF.

\begin{table}[h]
\centering
\caption{Multi-class datasets: Friedman's average ranking and $p$-values obtained through Holm post-hoc test using AutoEn as control method}
\label{tab:fiendman_testall_Holms_multiclass}
\resizebox{0.45\columnwidth}{!}{%
\begin{tabular}{ccc}
\hline
\textbf{Methods} & \textbf{Av. Ranking} & \textbf{$p$-values} \\ \hline
\textbf{AutoEn} & 2.25 & - \\ 
AutoSkl\_4h & 2.7083 & 0.6818 \\ 
\textbf{AutoEn\_ec} & 2.9167 & 0.5509 \\ 
AutoSkl\_1h & 3.6667 & 0.2051 \\ \
BestV\_ML & 4.0833 & 0.1010 \\ 
tuned\_RF & 5.75 & \textbf{0.0017} \\
RF & 6.7083 & \textbf{0.000067} \\ 
AutoW\_1h & 8.375 & \textbf{0} \\ 
AutoW\_4h & 8.5417 & \textbf{0} \\ \hline
\end{tabular}}
\end{table}

From Tables \ref{tab:fiendman_testall_Holms_binary}  and \ref{tab:fiendman_testall_Holms_multiclass}, it is possible to observe that despite the Friedman test provides a ranking of the methods evaluated, there were no statistical differences among some of them, particularly between the AutoML methods. Therefore, to better assess the performance of AutoEn and AutoEn\_ec, we introduce the following analyses. 

\begin{figure*}[!h]
 \centering
 \subfloat[AutoEn]{\label{autens_vs_all_in_binary} \includegraphics[width=5.9cm]{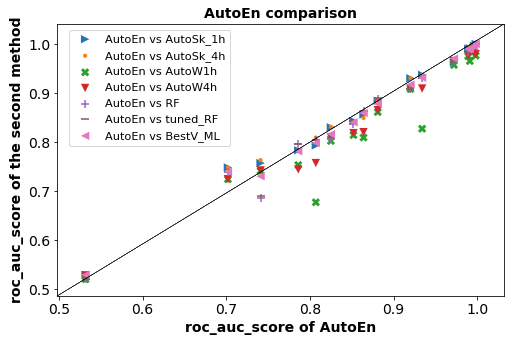}}
 \subfloat[AutoEn\_ec]{\label{autens_ec_vs_all_in_binary} \includegraphics[width=5.9cm]{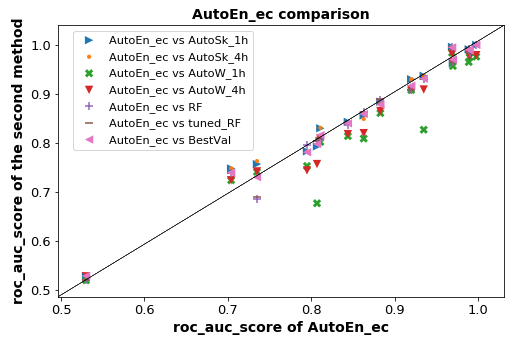}}
 \caption{roc\_auc\_score over 16 \textit{binary} datasets. Points
\textit{below} the $y = x$ line correspond to datasets for which our method
performs better than a comparison algorithm}
\end{figure*}

In Figures \ref{autens_vs_all_in_binary} and \ref{autens_ec_vs_all_in_binary}, each point compares AutoEn and AutoEn\_ec to a second method on binary problems, respectively. In both Figures, the x-axis position of the points is the roc\_auc\_score of our method on binary datasets. The y-axis position represents the performance metric of the comparison  algorithms. Points below the y=x line correspond to datasets for which our method performs better than a second method.

\begin{figure*}[h]
 \centering
 \subfloat[AutoEn]{\label{autens_vs_all_in_multiclass} \includegraphics[width=5.9cm]{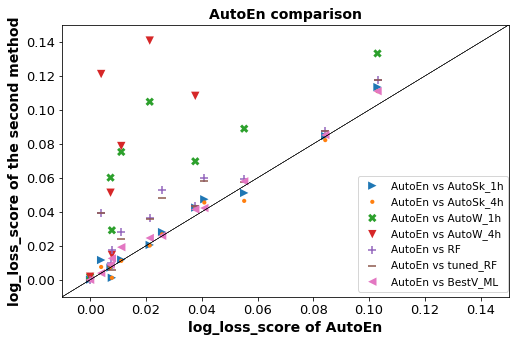}}
 \subfloat[AutoEn\_ec]{\label{autens_ec_vs_all_in_multiclass} \includegraphics[width=5.9cm]{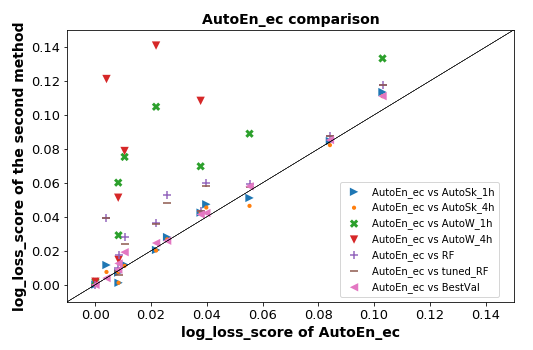}}
 \caption{log\_loss\_score over 12 \textit{multi-class} datasets. Points
\textit{above} the $y = x$ line correspond to datasets for which our method
performs better than a comparison algorithm}
\end{figure*}

Similarly, Figures \ref{autens_vs_all_in_multiclass} and \ref{autens_ec_vs_all_in_multiclass} compare AutoEn and AutoEn\_ec to a second method in multi-class datasets. Note that log\_loss\_score is a minimisation metric; therefore, points above y=x are datasets wherein AutoEn and AutoEn\_ec have a higher performance than a second method. For this latter case, the values are normalised between 0 and 1, and outliers were discarded. The reason to normalise the values is that, as was shown in Table \ref{Table:1-4_hours_results}, there are multi-class problems wherein some methods obtained values greater than 1.



In a general way, it can be observed for binary datasets, AutoEn and AutoEn\_ec have in the majority of the cases similar results to the comparison methods. On the other hand, in multi-class problems, our two approaches have a performance generally better than the other methods.

\begin{figure}[h]
\caption{AutoEn execution times in binary datasets under its default and economy modes}
\label{autens_time_in_binary}
\centering
\includegraphics[scale=0.25]{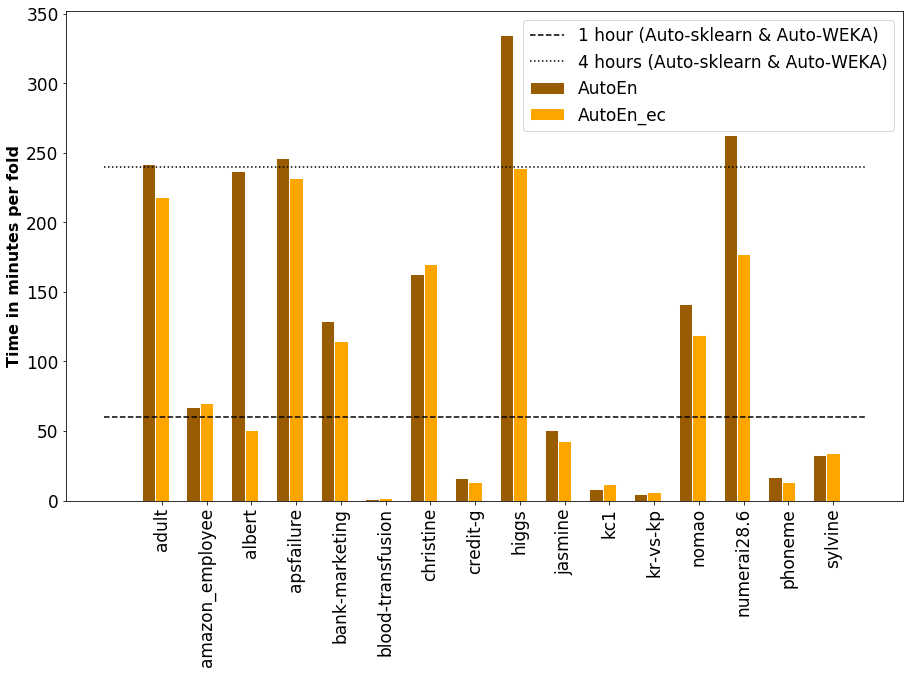}
\end{figure}

Finally, as the computational cost is also a relevant factor in AutoML, Figure \ref{autens_time_in_binary} shows the execution time per fold that AutoEn took to make classifications on binary datasets. The Figure also plots the execution time of its economy mode. Figure \ref{autens_time_in_binary} also has straight lines that represent the time thresholds of the two optimisation execution times associated with Auto-sklearn and Auto-WEKA. Thus, the purpose is to check whether the execution times of AutoEn can be in-between, below or beyond these two thresholds.

\begin{figure}[h]
\caption{AutoEn execution times in multi-class datasets under its default and economy modes}
\label{autens_time_in_multiclass}
\centering
\includegraphics[scale=0.25]{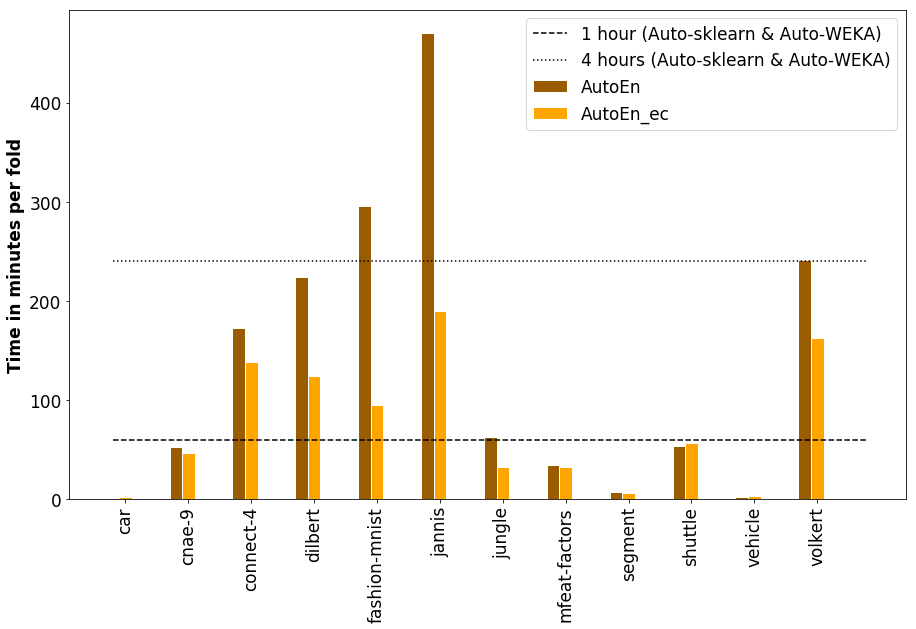}
\end{figure}

In Figure \ref{autens_time_in_binary}, the execution time of AutoEn in its default mode has three different behaviours. First, in \textit{blood-transfusion}, \textit{credit-g}, \textit{jasmine}, \textit{kc1}, \textit{kr-vs-kp}, \textit{phoneme} and \textit{sylvine} datasets, AutoEn spends less than 60 minutes per fold that is the shortest execution time allocated for Auto-sklearn and Auto-WEKA. Secondly, for \textit{adult}, \textit{amazon\_employee}, \textit{albert}, \textit{bank-marketing}, \textit{christine} and \textit{nomao} datasets, its execution time is in-between the the two thresholds. Finally, for the remaining datasets (\textit{apsfailure}, \textit{higgs}, \textit{numerai28.6}), AutoEn takes more than 4 hours of time. With resgard to AutoEn\_ec, its execution time is below 4 hours in all binary datasets, and as it was observed in Figure \ref{autens_ec_vs_all_in_binary}, its performance remains quite similar to AutoEn.

Figure \ref{autens_time_in_multiclass} summarises AutoEn and AutoEn\_ec execution times per fold of in multi-class datasets. Besides, it also shows the 2 time thresholds of the two optimisation execution times carried out by Auto-sklearn and Auto-WEKA. Significant reductions of time can be seen in \textit{dilbert}, \textit{fashion-mnist}, \textit{jannis} and \textit{volkert} datasets, without affecting the performance as was observed in Figure \ref{autens_ec_vs_all_in_multiclass}.

\subsection{Results in Traffic Forecasting}

This section presents and analyses the results obtained by AutoEn in TF. Specifically, we aim to compare the competitiveness and the significance of AutoEn with regard to a general-purpose AutoML method (Auto-sklearn) that is state-of-the-art in TF problems.

Table \ref{results_AutoEn_in_TF} shows the mean $log\_loss\_score$ values of our AutoML method, the AutoML competitors (AutoSk\_15m, AutoSk\_60m, AutoSk\_150m), and the baseline methods on the test phase. The best result for each dataset is highlighted in bold-face, and note that the lower the loss value (log\_loss), the better. Observing Table \ref{results_AutoEn_in_TF}, we can point out the following:

\begin{table}[h]
\centering
\caption{Mean $log\_loss$ (values obtained by AutoEn, the three ET of Auto-sklearn and the two baseline methods. Values highlighted in bold are the highest performance obtained by any of the methods on every dataset}
\label{results_AutoEn_in_TF}
\resizebox{\columnwidth}{!}{%
\begin{tabular}{cccccccc}
\hline
\textbf{Type} & \textbf{dataset} & \textbf{AutoEn} & \textbf{AutoSk\_15m} & \textbf{AutoSk\_60m} & \textbf{AutoSk\_150m} & \textbf{BestV\_ML} & \textbf{RF} \\ \hline
\multirow{10}{*}{Freeway} & T\_CD\_5m & \textbf{0.220} & 0.229 & 0.225 & 0.223 & 0.244 & 0.233 \\  
 & T\_CD\_15m & \textbf{0.340} & 0.360 & 0.346 & 0.348 & 0.371 & 0.366 \\
 & T\_CD\_30m & \textbf{0.381} & 0.398 & 0.389 & 0.391 & 0.417 & 0.442 \\
 & T\_CD\_45m & \textbf{0.393} & 0.438 & 0.429 & 0.420 & 0.439 & 0.472 \\ 
 & T\_CD\_60m & \textbf{0.401} & 0.499 & 0.454 & 0.453 & 0.424 & 0.489 \\ 
 & TS\_CD\_5m & 0.128 & 0.117 & \textbf{0.114} & 0.116 & 0.142 & 0.133 \\
 & TS\_CD\_15m & 0.166 & \textbf{0.163} & 0.164 & 0.169 & 0.187 & 0.185 \\ 
 & TS\_CD\_30m & \textbf{0.186} & 0.193 & 0.192 & 0.197 & 0.212 & 0.218 \\
 & TS\_CD\_45m & \textbf{0.180} & 0.195 & 0.182 & 0.182 & 0.199 & 0.217 \\
 & TS\_CD\_60m & \textbf{0.182} & 0.218 & 0.197 & 0.204 & 0.185 & 0.229 \\ \hline
\multirow{8}{*}{Urban} & T\_CD\_15m & 0.530 & 0.502 & \textbf{0.500} & 0.503 & 0.537 & 0.539 \\ 
 & T\_CD\_30m & 0.538 & 0.516 & 0.510 & \textbf{0.505} & 0.536 & 0.568 \\
 & T\_CD\_45m & 0.530 & 0.513 & 0.512 & \textbf{0.495} & 0.537 & 0.571 \\
 & T\_CD\_60m & 0.534 & 0.508 & \textbf{0.500} & 0.508 & 0.553 & 0.569 \\
 & TS\_CD\_15m & 0.406 & 0.390 & \textbf{0.387} & 0.389 & 0.412 & 0.420 \\
 & TS\_CD\_30m & \textbf{0.429} & 0.435 & 0.434 & 0.437 & 0.468 & 0.461 \\
 & TS\_CD\_45m & \textbf{0.439} & 0.447 & 0.453 & 0.452 & 0.494 & 0.482 \\
 & TS\_CD\_60m & \textbf{0.435} & 0.437 & 0.438 & 0.437 & 0.457 & 0.485 \\ \hline
\multicolumn{2}{c}{\textbf{Wins}} & \textit{11} & \textit{1} & \textit{4} & \textit{2} & \textit{0} & \textit{0} \\ \hline
\end{tabular}}
\end{table}

\begin{itemize}
    \item As a general overview of results, AutoEn is the most competitive learning approach. This can be seen in the wins distribution as follows: (11) AutoEn, (4) AutoSk\_60m, (2) AutoSk\_150m, and (1) AutoSk\_15m. Moreover, AutoEn and the three ETs of AutoSk are better than RF over all datasets; while they are better than BestV\_ML in almost all datasets.
    
    \item In freeway datasets, AutoEn is the better approach, especially in time horizons of predictions that range from 30 to 60 minutes in datasets with temporal-spatial (TS) traffic data. On the other hand, AutoEn is the better approach in all datasets that have only temporal (T) traffic data. Besides, AutoSk\_15s and AutoSk\_60m are more competitive in shorter time horizons (5 and 15 minutes) of freeway datasets with TS traffic data. In this sense, AutoEn achieves to offer competitive results over long-term time horizons that were identified as a drawback of other AutoML approaches in previous resarch \cite{Angarita2018,Angarita2020,Angarita-Zapata2020}.
    
    \item In urban datasets, AutoEn is more competitive in datasets that contain temporal-spatial (TS) data. Contrary, its performance is just behind Auto-sklearn in datasets with only temporal (T) traffic data.
    
    \item As it was previously discussed in results of the general-purpose domain showed in Section \ref{Results}, longer ETs allocated for the optimisation do not guarantee better performance. This assumption work properly in some datasets (freeway: \textit{T\_CD\_5m}, \textit{T\_CD\_45m}, \textit{T\_CD\_60m}; urban: \textit{T\_CD\_30m}, \textit{T\_CD\_45m}). However, in some other cases, longer ETs end up with similar results to the ones obtained with shorter ETs (freeway: \textit{TS\_CD\_5m}, \textit{TS\_CD\_30m}; urban: \textit{T\_CD\_15m}, \textit{TS\_CD\_60m}); or in the worst of the cases, they tend to decrease the performance of Auto-sklearn (freeway: \textit{TS\_CD\_15m}, \textit{TS\_CD\_30m}; urban: \textit{TS\_CD\_15m}, \textit{TS\_CD\_30m}).
    
\end{itemize}

To assess whether the differences in performance observed in Table \ref{results_AutoEn_in_TF} are significant or not, we used the same non-parametric statistical tests mentioned before. Table \ref{tab:fiendman_testall_Holms_TF} exposes the test outcomes and the $p$-values lower than 0.05 are shown in bold. In this case, AutoEn is the method in the first position of the ranking. However, it is only statistical better than RF and BestV\_ML.

\begin{table}[h]
\centering
\caption{Friedman's average ranking and $p$-values obtained through Holm post-hoc test using AutoEn as control method}
\label{tab:fiendman_testall_Holms_TF}
\resizebox{0.4\columnwidth}{!}{%
\begin{tabular}{ccc}
\hline
\textbf{Methods} & \textbf{Av. Ranking} & \textbf{$p$-values} \\ \hline
\textbf{AutoEn} & 2.1667 & - \\ 
AutoSk\_60m & 2.3333 & 0.9520 \\ 
AutoSk\_150m & 3.6111 & 0.9520 \\ \
AutoSk\_15m & 3.3333 & 0.1841\\ 
BestV\_ML & 4.9444 & \textbf{0} \\
RF & 5.6111 & \textbf{0} \\ \hline
\end{tabular}}
\end{table}

\begin{figure}[h]
\caption{Log loss score over 18 traffic forecasting datasets. Points \textit{above} the $y = x$ line correspond to datasets for which our method performs better than a comparison algorithm}
\label{AutoEc_comparison_in_TF}
\centering
\includegraphics[scale=0.4]{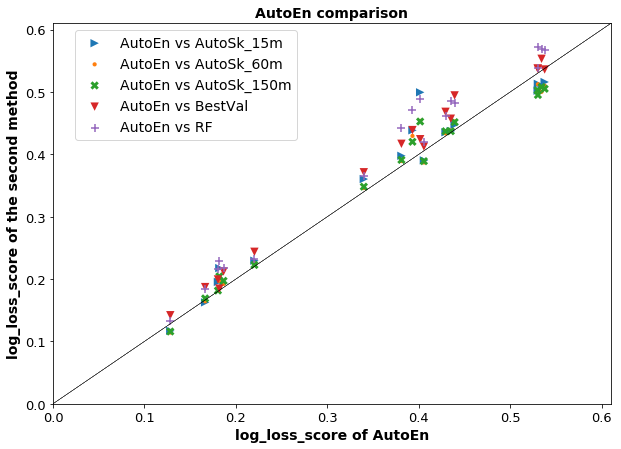}
\end{figure}

From Tables \ref{tab:fiendman_testall_Holms_TF}, it is possible to observe that despite the Friedman's test provides a ranking of the methods evaluated, there were no statistical differences among some of them, particularly between the AutoML methods. Therefore, to better assess the performance, we introduce the following analyses of Figure \ref{AutoEc_comparison_in_TF}. Each point compares AutoEn to a second method on the multi-class datasets. The x-axis position of the points is the log\_loss\_score of our method on, while the y-axis position represents the performance metric of the comparison algorithms. Points above the y=x line correspond to datasets for which AutoEn performs better than a second method. From Figure \ref{AutoEc_comparison_in_TF} it can be observed that AutoEn has in most of the cases better results than the comparison methods.

In summary,  AutoEn can overcome the optimisation and meta-learning issues of AutoML in TF. A simple strategy based on the construction of multi-classifiers systems achieves better results in TF problems without pre-defined time budgets for the optimisation process and without being limited by the representativeness of meta-learning. Although AutoSk\_60m and AutoSk\_150m achieved better results than AutoEn in some particular cases, non-expert ML user must test Auto-sklearn multiple times using different ETs to find these competitive results.




\section{Conclusions}
\label{Conclusions}

In this paper, we introduced AutoEn, a new AutoML method for supervised problems that has a search strategy supported by the construction of ensembles of multiple classifiers. AutoEn was tested against Auto-sklearn and Auto-WEKA, two state-of-the-art AutoML methods, in multiple binary and multi-class supervised problems of a well-established AutoML benchmark, which have been used in TF previously. Besides, we test AutoEn against Auto-sklearn in TF, considering various multi-class TF problems. The main conclusions drawn from the results are:

\begin{itemize}
    \item AutoEn can be better and more competitive than state-of-the-art AutoML approaches in supervised TF problems. Notably, it is less prone to overfitting than Auto-sklearn because its ensemble strategy makes more robust predictions as multiple classifiers are considered at the same time. Besides, AutoEn can better adapt to different time horizons, specifically, in long-term time horizons that are a crucial element for transportation users when are managing traffic flows.
    
    \item Finally, it is relevant to note that in the TF domain, transportation users non-experts in ML could use other AutoML methods focused on optimisation of individual pipelines and achieve satisfactory results. However, as the core of other AutoML strategies is based on optimisation, the user should test multiple times different time budgets to deliver competitive results. This is a task that involves human effort and time costs, which does not have clear guidelines to decide the best ET depending on the size of the dataset at hand. Therefore, AutoEn arises a promising approach that overcomes these limitations while it can supply the demands of non-expert ML users in TF problems.
    
    \item Ensembles of multi-classifiers can be more adaptable to datasets of different sizes, which leads to optimising the computational efficiency of AutoML in TF and other problem domains. On the one hand, high-performance solutions can be achieved for small- and medium-size datasets faster than traditional optimisation approaches (e.g., Bayesian optimisation, evolutionary programming). On the contrary, although the time consumption could increase for big datasets, the computation effort of the online phase is focused on finding competitive combinations of multi-classifiers. Moreover, AutoEn's approach is highly parallelisable compared to generating and fine-tuning individual pipelines that are even more costly to be evaluated in big datasets.
    
    \item The online phase of AutoML should be focused on building ensembles rather than individual ML pipelines. Thus, computational effort should be directed to generate high-performance pipelines in the offline period of AutoML, which later contribute to producing competitive and fast ensembles during the online stage of AutoML. These ensembles are less prone to overfitting and are not exposed to the drawbacks of meta-learning and meta-features, as is supported by other general-purpose AutoML methods like Auto-Gluon \cite{Erickson2020}. Although AutoEn's ensemble approach can also be optimised, its hyperparameters are a reduced set of values. Such a characteristic of AutoEn constitutes a much less complicated search space compared to the online optimisation of pipelines with a higher degree of hyperparameters.

\end{itemize}


\bibliographystyle{elsarticle-num}

\begin{thebibliography}{10}
\expandafter\ifx\csname url\endcsname\relax
  \def\url#1{\texttt{#1}}\fi
\expandafter\ifx\csname urlprefix\endcsname\relax\def\urlprefix{URL }\fi
\expandafter\ifx\csname href\endcsname\relax
  \def\href#1#2{#2} \def\path#1{#1}\fi

\bibitem{Angarita_etal_2019}
J.~S. {Angarita-Zapata}, A.~D. {Masegosa}, I.~{Triguero}, A taxonomy of traffic
  forecasting regression problems from a supervised learning perspective, IEEE
  Access.

\bibitem{automl_book2018}
F.~Hutter, L.~Kotthoff, J.~Vanschoren (Eds.), Automated Machine Learning:
  Methods, Systems, Challenges, Springer, 2018.

\bibitem{Song2019}
H.~Song, I.~Triguero, E.~{\"O}zcan, {A review on the self and dual interactions
  between machine learning and optimisation}, Progress in Artificial
  Intelligence (2019) 1--23.

\bibitem{Luo2016}
G.~Luo, A review of automatic selection methods for machine learning algorithms
  and hyper-parameter values, Network Modeling Analysis in Health Informatics
  and Bioinformatics 5~(1) (2016) 18.

\bibitem{Yao2018}
Q.~Yao, M.~Wang, Y.~Chen, W.~Dai, H.~Yi-Qi, L.~Yu-Feng, T.~Wei-Wei, Y.~Qiang,
  Y.~Yang, {Taking Human out of Learning Applications: A Survey on Automated
  Machine Learning}, CoRR.

\bibitem{Zoller2019}
M.-A. Z{\"{o}}ller, M.~F. Huber, {Survey on Automated Machine Learning}, arXiv
  preprint arXiv:1904.12054\href {http://arxiv.org/abs/1904.12054}
  {\path{arXiv:1904.12054}}.

\bibitem{VLAHOGIANNI201514}
E.~I. Vlahogianni,
  \href{http://www.sciencedirect.com/science/article/pii/S0968090X15000959}{Optimization
  of traffic forecasting: Intelligent surrogate modeling}, Transportation
  Research Part C: Emerging Technologies 55 (2015) 14 -- 23, engineering and
  Applied Sciences Optimization (OPT-i) - Professor Matthew G. Karlaftis
  Memorial Issue.
\newline\urlprefix\url{http://www.sciencedirect.com/science/article/pii/S0968090X15000959}

\bibitem{Angarita2018}
J.~S. Angarita-Zapata, I.~Triguero, A.~D. Masegosa, A preliminary study on
  automatic algorithm selection for short-term traffic forecasting, in:
  J.~Del~Ser, E.~Osaba, M.~N. Bilbao, J.~J. Sanchez-Medina, M.~Vecchio, X.-S.
  Yang (Eds.), Intelligent Distributed Computing XII, Springer International
  Publishing, Cham, 2018, pp. 204--214.

\bibitem{Angarita2020}
J.~S. Angarita-Zapata, A.~D. Masegosa, I.~Triguero, General-purpose automated
  machine learning for transportation: A case study of auto-sklearn for traffic
  forecasting, in: Proceeding of the 18th International Conference on
  Information Processing and Management of Uncertainty in Knowledge-Based
  Systems, Springer International Publishing, Cham, 2020.

\bibitem{Angarita-Zapata2020}
J.~S. Angarita-Zapata, A.~D. Masegosa, I.~Triguero, Evaluating Automated
  Machine Learning on Supervised Regression Traffic Forecasting Problems,
  Springer International Publishing, Cham, 2020, pp. 187--204.

\bibitem{Thornton2013}
C.~Thornton, F.~Hutter, H.~H. Hoos, K.~Leyton-Brown, Auto-weka: Combined
  selection and hyperparameter optimization of classification algorithms, in:
  Proceedings of the 19th International Conference on Knowledge Discovery and
  Data Mining, Association for Computing Machinery, 2013, p. 847–855.

\bibitem{Feurer2015}
M.~Feurer, A.~Klein, K.~Eggensperger, J.~Springenberg, M.~Blum, F.~Hutter,
  {Efficient and Robust Automated Machine Learning}, in: C.~Cortes, N.~D.
  Lawrence, D.~D. Lee, M.~Sugiyama, R.~Garnett (Eds.), Advances in Neural
  Information Processing Systems, Curran Associates, Inc., 2015, pp.
  2962--2970.

\bibitem{Gijsbers2019}
P.~Gijsbers, E.~LeDell, S.~Poirier, J.~Thomas, B.~Bischl, J.~Vanschoren, An
  open source automl benchmark, CoRRWork presented at AutoML Workshop at
  International Conference on Machine Learning 2019.

\bibitem{GALAR2011}
M.~Galar, A.~Fernandez, E.~Barrenechea, H.~Bustince, F.~Herrera, An overview of
  ensemble methods for binary classifiers in multi-class problems: Experimental
  study on one-vs-one and one-vs-all schemes, Pattern Recognition 44~(8) (2011)
  1761 -- 1776.

\bibitem{Bishop2006}
C.~M. Bishop, Pattern Recognition and Machine Learning (Information Science and
  Statistics), Springer-Verlag, Berlin, Heidelberg, 2006.

\bibitem{Garcia2014}
S.~Garcia, J.~Luengo, F.~Herrera, Data preprocessing in data mining,
  \textit{Springer.} Cham, Switzerland, 2015.

\bibitem{triguero19}
I.~Triguero, D.~Garc{\'\i}a-Gil, J.~Maillo, J.~Luengo, S.~Garc{\'\i}a,
  F.~Herrera, Transforming big data into smart data: An insight on the use of
  the k-nearest neighbors algorithm to obtain quality data, \textit{Wiley
  Interdisciplinary Reviews: Data Mining and Knowledge Discovery} 9~(2) (2019)
  e1289.

\bibitem{Kanter2015}
J.~M. {Kanter}, K.~{Veeramachaneni}, Deep feature synthesis: Towards automating
  data science endeavors, in: 2015 IEEE International Conference on Data
  Science and Advanced Analytics, 2015, pp. 1--10.

\bibitem{Katz2016}
G.~{Katz}, E.~C.~R. {Shin}, D.~{Song}, Explorekit: Automatic feature generation
  and selection, in: 2016 IEEE 16th International Conference on Data Mining,
  2016, pp. 979--984.

\bibitem{Nargesian2017}
F.~Nargesian, H.~Samulowitz, U.~Khurana, E.~B. Khalil, D.~Turaga, Learning
  feature engineering for classification, in: Proceedings of the 26th
  International Joint Conference on Artificial Intelligence, AAAI Press, 2017,
  pp. 2529--2535.

\bibitem{Bergstra2011}
J.~Bergstra, R.~Bardenet, Y.~Bengio, B.~K{\'e}gl, Algorithms for
  hyper-parameter optimization, in: Proceedings of the 24th International
  Conference on Neural Information Processing Systems, Curran Associates Inc.,
  USA, 2011, pp. 2546--2554.

\bibitem{Hutter2011}
F.~Hutter, H.~H. Hoos, K.~Leyton-Brown, {Sequential Model-Based Optimization
  for General Algorithm Configuration}, in: C.~A.~C. Coello (Ed.), Learning and
  Intelligent Optimization, Springer Berlin Heidelberg, Berlin, Heidelberg,
  2011, pp. 507--523.

\bibitem{Claesen2014}
M.~Claesen, J.~Simm, D.~Popovic, Y.~Moreau, B.~D. Moor, Easy hyperparameter
  search using optunity, CoRR.

\bibitem{Olson2016}
R.~S. Olson, N.~Bartley, R.~J. Urbanowicz, J.~H. Moore, {Evaluation of a
  Tree-based Pipeline Optimization Tool for Automating Data Science}, in:
  Proceedings of the Genetic and Evolutionary Computation Conference 2016,
  2016, pp. 485--492.

\bibitem{vanschoren2019}
J.~Vanschoren, Meta-learning, in: Hutter et~al.  \cite{automl_book2018}, pp.
  39--68.

\bibitem{Erickson2020}
N.~Erickson, J.~Mueller, A.~Shirkov, H.~Zhang, P.~Larroy, M.~Li, A.~Smola,
  Autogluon-tabular: Robust and accurate automl for structured data, arXiv
  preprint arXiv:2003.06505.

\bibitem{Caruana2004}
R.~Caruana, A.~Niculescu-Mizil, G.~Crew, A.~Ksikes, Ensemble selection from
  libraries of models, in: Proceedings of the Twenty-First International
  Conference on Machine Learning, Association for Computing Machinery, New
  York, NY, USA, 2004, p.~18.

\bibitem{Isabelle2016}
I.~Guyon, I.~Chaabane, H.~J. Escalante, S.~Escalera, D.~Jajetic, J.~R. Lloyd,
  N.~Macià, B.~Ray, L.~Romaszko, M.~Sebag, A.~R. Statnikov, S.~Treguer,
  E.~Viegas, A brief review of the chalearn automl challenge: Any-time
  any-dataset learning without human intervention, in: Proceedings of the
  Workshop on Automatic Machine Learning, PMLR, New York, USA, 2016, pp.
  21--30.

\bibitem{bischl2017}
B.~Bischl, G.~Casalicchio, M.~Feurer, F.~Hutter, M.~Lang, R.~G. Mantovani,
  J.~N. van Rijn, J.~Vanschoren, Openml benchmarking suites (2017).
\newblock \href {http://arxiv.org/abs/1708.03731} {\path{arXiv:1708.03731}}.

\bibitem{Vanschoren2014}
J.~Vanschoren, J.~N. van Rijn, B.~Bischl, L.~Torgo, Openml: Networked science
  in machine learning, SIGKDD Explor. Newsl. 15~(2) (2014) 49–60.

\bibitem{Skycomp2009}
I.~B.~M. Skycomp, {Major High- way Performance Ratings and Bottleneck
  Inventory}, Maryland State Highway Administration, the Baltimore Metropolitan
  Council and Maryland Transportation Authority, State of Maryland, 2009.

\bibitem{GARCIA2010}
S.~Garcia, A.~Fernandez, J.~Luengo, F.~Herrera, Advanced nonparametric tests
  for multiple comparisons in the design of experiments in computational
  intelligence and data mining: Experimental analysis of power, Information
  Sciences 180~(10) (2010) 2044 -- 2064.

\bibitem{Falkne2018}
S.~Falkner, A.~Klein, F.~Hutter, Bohb: Robust and efficient hyperparameter
  optimization at scale, CoRR.

\end{thebibliography}

\end{document}